\documentclass[letterpaper, 10 pt, conference]{ieeeconf}  

\IEEEoverridecommandlockouts                              

\usepackage{graphicx} 
\usepackage{amsmath} 
\usepackage{float}

\title{\LARGE \bf
Applying Neural Monte Carlo Tree Search to Unsignalized Multi-intersection Scheduling for Autonomous Vehicles
}

\author{Yucheng Shi$^{1}$, Wenlong Wang$^{2}$, Xiaowen Tao$^{3}$, Ivana Dusparic$^{4}$ and Vinny Cahill$^{5}$
\thanks{*This work was supported by Science Foundation of Ireland}
\thanks{$^{1}$Yucheng Shi, School of Computer Science and Statistics, Trinity College Dublin, Ireland
        {\tt\small shiy2@tcd.ie}}%
\thanks{$^{2}$Wenlong Wang, School of Computer Science and Statistics, Trinity College Dublin, Ireland
        {\tt\small wangw1@tcd.ie}}%
\thanks{$^{3}$Xiaowen Tao, School of Computer Science and Statistics, Trinity College Dublin, Ireland
        {\tt\small taox@tcd.ie}}%
\thanks{$^{4}$Ivana Dusparic, School of Computer Science and Statistics, Trinity College Dublin, Ireland
        {\tt\small ivana.dusparic@tcd.ie}}%
\thanks{$^{5}$Vinny Cahill,School of Computer Science and Statistics, Trinity College Dublin, Ireland
        {\tt\small vjcahill@tcd.ie}}%
}

\begin{document}

\maketitle
\thispagestyle{empty}
\pagestyle{empty}

\begin{abstract}

Dynamic scheduling of access to shared resources by autonomous systems is a challenging problem, characterized as being NP-hard. The complexity of this task leads to a combinatorial explosion of possibilities in highly dynamic systems where arriving requests must be continuously scheduled subject to strong safety and time constraints. An example of such a system is an unsignalized intersection, where automated vehicles' access to potential conflict zones must be dynamically scheduled. In this paper, we apply Neural Monte Carlo Tree Search (NMCTS) to the challenging task of scheduling platoons of vehicles crossing unsignalized intersections. Crucially, we introduce a transformation model that maps successive sequences of potentially conflicting road-space reservation requests from platoons of vehicles into a series of board-game-like problems and use NMCTS to search for solutions representing optimal road-space allocation schedules in the context of past allocations. To optimize search, we incorporate a prioritized re-sampling method with parallel NMCTS (PNMCTS) to improve the quality of training data. To optimize training, a curriculum learning strategy is used to train the agent to schedule progressively more complex boards culminating in overlapping boards that represent busy intersections. In a busy single four-way unsignalized intersection simulation, PNMCTS solved 95\% of unseen scenarios, reducing crossing time by 43\% in light and 52\% in heavy traffic versus first-in, first-out control. In a 3x3 multi-intersection network, the proposed method maintained free-flow in light traffic when all intersections are under control of PNMCTS and outperformed state-of-the-art RL-based traffic-light controllers in average travel time by 74.5\%  and total throughput by 16\% in heavy traffic. 

\end{abstract}

\section{INTRODUCTION}

Dynamic scheduling of access to shared resources by autonomous systems, such as allocating road space to the flows of connected automated vehicles (CAVs) crossing an unsignalized junction, is a challenging problem due to safety and timeliness constraints. This is essentially a combinatorial optimisation problem closely related to the dynamic job shop scheduling problem, which has been proven to be NP-hard, signifying that its computational complexities increase exponentially as the problem scales \cite{Garey1976ComplexityScheduling}. To tackle this problem, we apply a reinforcement learning (RL)-based approach to identifying safe and efficient schedules. The need for (near) real-time decision-making is crucial for scheduling steady streams of incoming vehicles and their road-space reservation requests. Furthermore, since the arrival of vehicles is continuous, while a previous schedule is still being executed new vehicles may arrive that must be seamlessly integrated into the existing plan, such that there are no resource conflicts between the new vehicles and those previously scheduled. This adds complexity as the scheduling algorithm must manage a continuous vehicle flow while accommodating new arrivals seamlessly. For CAVs, it requires making safe, time-bound scheduling decisions that minimize crossing time for arriving vehicles or platoons. 

Our approach is motivated by the recent success of deep reinforcement learning (DRL) in demonstrating exceptional performance in tackling a variety of scheduling and optimisation challenges,  ranging from complex combinatorial optimisation tasks \cite{mazyavkina2021reinforcement} to intelligent manufacturing systems \cite{shiue2018real}. Many studies \cite{shiue2018real,zhang2021digital} have used Q-learning for optimizing dynamic scheduling in discrete, finite state spaces. However, since Q-learning relies on a behavior policy for experience collection, its effectiveness is highly dependent on adequate exploration. This poses challenges in environments with large, sparse state spaces.

To address this limitation, neural Monte-Carlo tree search (NMCTS), which combines the advantages of  Monte Carlo tree search (MCTS) \cite{silver2018general} and policy-based RL \cite{sutton1999policy}, can be leveraged. AlphaZero \cite{schrittwieser2020mastering} is the most prominent example of such a combination, mastering diverse board games without any prior knowledge provided by humans.
An NMCTS-based method allows the model to better explore the environment using upper confidence bound-based (UCB) exploration that prioritises actions that are under-explored or show high potential, thereby enhancing the algorithm's focus on crucial states that could have significant impact. Moreover, deep neural networks (DNN) work as a memory unit to guide the search, focusing exploration on the most promising moves based on previously acquired knowledge via the self-play technique \cite{liu2021sharp}. A detailed explanation of MCTS can be found in \cite{silver2018general}.

Our first contribution lies in building a transformation model that abstracts the real-world intersection dynamics into a schedulable 'board game'. This allows us to recast the challenges of platoon management within the context of NMCTS. The objective then becomes straightforward: minimizing the time required to cross the intersection by delaying platoon entry such that no two platoons are simultaneously present in a collision zone, thereby ensuring collision-free scheduling. We then propose a parallel approach to NMCTS (PNMCTS) that utilises four training strategies: Firstly, we advance the tree search by adopting a parallelised multi-process approach, allowing multiple tree searches and simulations to roll out simultaneously, such that every process can start from a distinct state and operates on an independent tree. Once all processes conclude their search, the trajectories are aggregated. Secondly, the system catalogs the optimal trajectories for each resolved scenario, periodically reintroducing them to the training pool, serving to enhance the reward signal. Thirdly, we add entropy regulation to the policy loss function. This technique encourages more expansive exploration throughout the training phase, acting as a countermeasure to the constraints posed by the pruned tree search. Lastly, to tackle the issue of continuous vehicle flow, we introduce a curriculum learning strategy \cite{narvekar2020curriculum} with a short-path MCTS roll out into the training process. This allows the policy to adapt to handle increasingly complex situations encompassing residual platoons scheduled by prior decision making. 

In the SUMO \cite{lopez2018microscopic} simulation environment, the PNMCTS method was tested on a single four-way three-lane unsignalized intersection, managing to effectively solve unseen, high-density traffic scenarios.  In a 3x3 grid network, PNMCTS maintained free-flow traffic under light traffic and outperformed state-of-the-art RL-based traffic-light controllers.

\section{Related Work}

\subsection{Individual unsignalized intersection management }
As CAVs can approach intersections from diverse directions and at different times, the optimal crossing order continually changes, presenting a challenge for determining the optimal sequence using brute-force methods. Typically, researchers have proposed reservation-based methods to solve this problem \cite{dresner2004multiagent}, where the autonomous vehicle reserves the road space that they need before they arrive. The FIFO strategy is the most straightforward reservation-based method and is adopted in \cite{malikopoulos2018decentralized},where the authors concluded that the FIFO queue imposed limitations, especially in heavy traffic. \cite{10397236} modeled intersections based on the inner collision points and utilised tree search approach to arrange crossing priority.

Several RL-based approaches \cite{10050400,tram2018learning} adopt a paradigm where individual vehicles serve as learning agents and employ self-organizing techniques to collectively achieve a shared objective at the system level. \cite{10050400} developed a framework utilizing the Twin Delayed Deep Deterministic Policy Gradient (TD3) algorithm, specifically designed to manage vehicle dynamics at a multi-lane single intersection. \cite{tram2018learning} proposed a deep Q-learning method that can schedule unsignalized intersection crossing by modeling the problem as a partially observable Markov decision process. Both works focus on single vehicle control and leverage a value-based RL method, which are distinctively different from our approach. Our proposed method is specifically designed to address heavy traffic conditions, which requires the capability to integrate newly arriving vehicles seamlessly without rescheduling already scheduled vehicles. Additionally, we evaluate our approach across a network of multiple intersections, an aspect not commonly addressed in the field of unsignalized intersection management.

\subsection{Multi-intersection traffic light control}

Fixed-time traffic control \cite{roess2004traffic} employs pre-set schedules for signal changes, irrespective of actual traffic conditions, which lacks the flexibility of adaptive systems.

Actuated traffic light control as implemented in SUMO \cite{lopez2018microscopic} switches to the next phase after detecting a sufficient time gap between successive vehicles. This allows for better distribution of green-time among phases and also modifies cycle duration in response to dynamic traffic conditions.

LIT (Light-intelligent) \cite{zheng2019diagnosing} is a DRL approach designed for the management of multiple intersections. A key feature of this method is its simple reward system, which emphasizes queue length reduction without considering the state of downstream traffic.

PressLight \cite{wei2019presslight} advances LIT in policy training by using the concept of traffic pressure as its reward. Traffic pressure, in this context, refers to the differential between the number of vehicles approaching an intersection and those leaving it. In multiple intersection scenarios, PressLight allows for a more fluid traffic management system, aiming to maintain a steady flow of vehicles through intersections and reduce overall travel time. The above methods also serve as benchmarks for comparison with our proposed PNMCTS approach in Section IV.

\section{PROPOSED FRAMEWORK}

\subsection{Traffic transformation model}
\begin{figure}[h]
    \centering
    \includegraphics[width=0.4\linewidth]{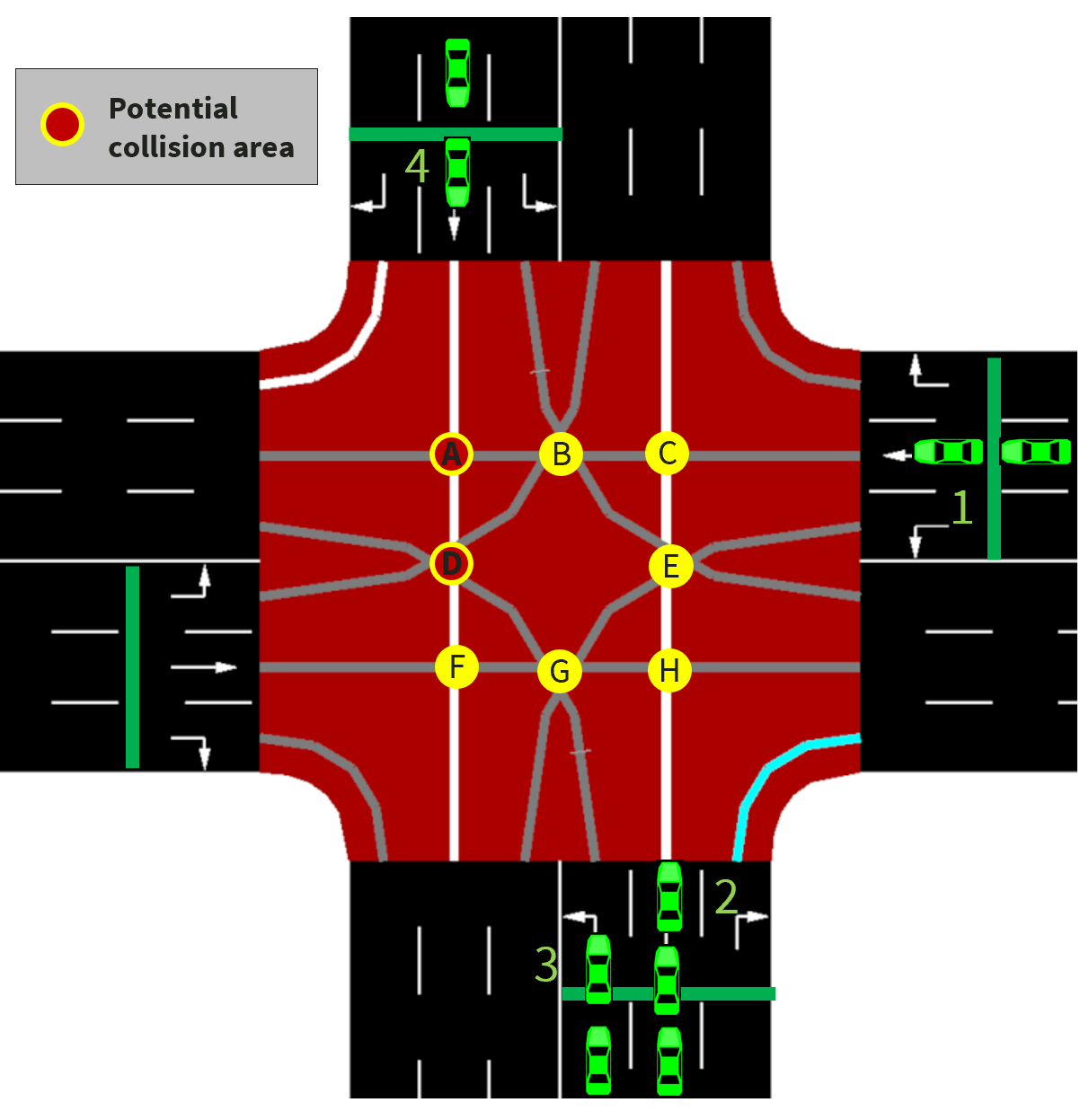}
    \includegraphics[width=0.5\linewidth]{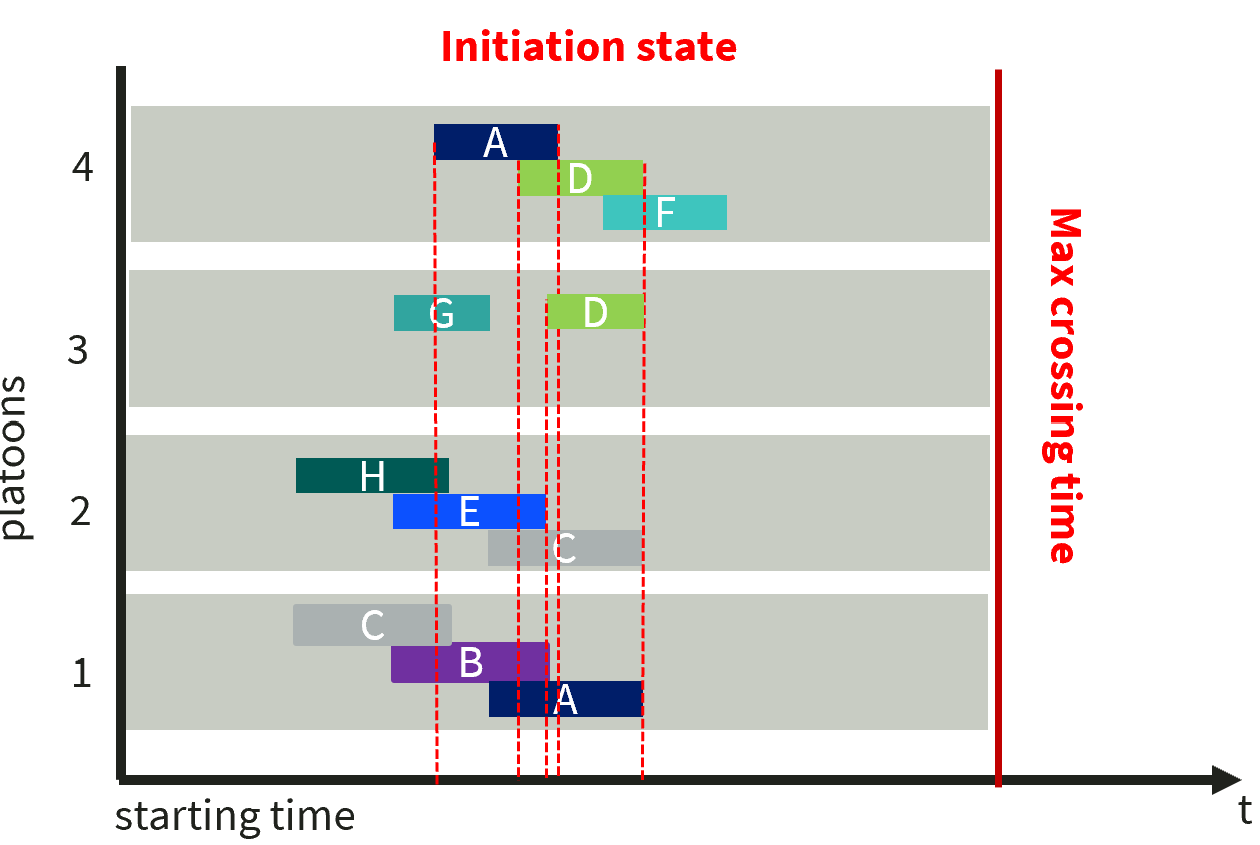}
   \caption{An intersection with four approaching platoons is illustrated on the left. The representative board corresponding to this scenario is given on the right. Entry and exit times for collision areas are shown on the time line (X-axis), while the Y-axis shows platoons numbers. As platoon 1 is moving forward, it will cross collision areas C$\rightarrow$ B $\rightarrow$ A in that order, with the occupancy time of the corresponding collision area sequence mapped to time blocks (shown stacked for each platoon for readability). The occupancy time of C, B, and A collision areas by platoon 1 overlap because the platoon is longer than the gap between the collision areas. We can observe from the illustration that, for example, without a scheduling intervention, platoons 1 and 4 will collide in collision area A, Platoon 3 and 4 will collide in collision area D, as they would occupy it simultaneously.}
    \label{fig:corresponding}
\end{figure}

In order to apply PNMCTS in the unsignalized intersection management scenario, we begin by considering how to schedule a fixed set of platoons crossing the junction. The first step is to apply a transformation model allowing us to map a sequence of resource requests corresponding to platoons of vehicles crossing the potential collision points in the intersection to a game board-like representation. In the intersection scenario, every approaching platoon comes with a pre-defined path to cross the intersection. Integral to this path are a set of collision areas (dots A-H in Fig. \ref{fig:corresponding} (left)), essentially, areas where vehicles' possible paths intersect within the junction to be crossed. Assuming that each incoming platoon crosses the intersection at the highest permissible speed, we can forecast an initial board configuration. This configuration will indicate the duration during which each platoon crosses each collision point in its crossing path. Such dynamics can be projected to a board represented as shown in Fig.  \ref{fig:corresponding} (right).

    


The initial board reflects a scheduling strategy in which all incoming platoons cross the intersection at their maximum allowed speed. However, given that platoon paths may not always be compatible, there is a significant probability of conflicts occurring at conflict areas, represented as overlapping blocks  on the board, e.g., platoons 3 and 4 conflict at D in Fig. \ref{fig:corresponding} (right).  To address this challenge, we apply a strategy that delays the time at which specific platoons enter the collision areas. It is important to note that any delays imposed on entering a collision zone will propagate backwards to (i.e., affect the time at which platoons can enter) previous collision zones and these adjustments can be represented as revised boards. By adjusting the schedule in this manner, we ensure that no two platoons are concurrently present in the same collision area, thereby preventing any collision from happening. In general, the problem involves finding mutually compatible sequences of blocks that enable all platoons to cross the junction in the shortest possible time considering an arbitrary set of platoons to be scheduled. This is addressed using PNMCTS. In the context of employing RL techniques for sequential decision-making, and leveraging PNMCTS to explore optimal scheduling strategies within this environment, we formalise the definitions of the states, actions, and rewards required by the RL paradigm as follows:

\subsubsection{State}

\begin{figure}[h]
  \centering
  \includegraphics[width=0.7\linewidth]{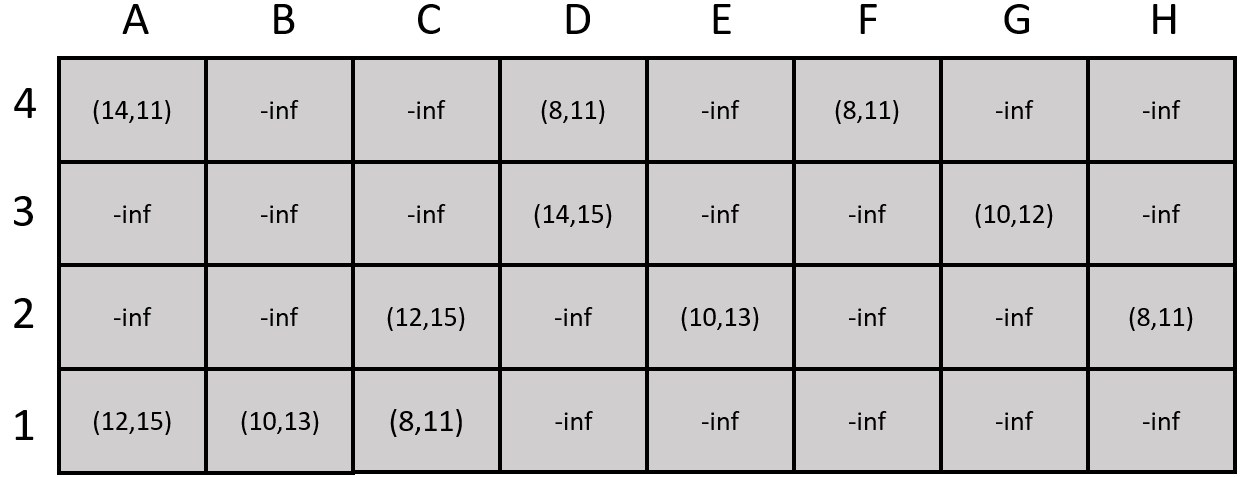}
  \caption{Grid representation of transformation board.}
  \label{fig:grid}
\end{figure}

The state is represented as a grid, shown in Fig. \ref{fig:grid}, dimensioned by the maximum number of crossing platoons \(N\) and the total number of collision areas \(M\). Each grid cell contains a tuple of two values: $\langle t_{entry},  t_{exit} \rangle$, which correspond to the \(N^{th}\) platoon's occupancy duration in the \(M^{th}\) collision area. The mathematical representation is given in (\ref{eq:state}). 

\begin{equation}
\label{eq:state}
\text S = \left\{\left\{\{t^{i,j}_{entry},t^{i,j}_{exit}\}_{i\in M}\right\}_{j\in N}\right\}
\end{equation}

If a collision area is not part of the platoon's crossing path, the tuple defaults to the placeholder value of $-\inf$. This state representation comprehensively summarises all the essential geometric features and the dynamics of the intersection, including platoon size, travel speed, chosen path, distance to the intersection and collision area distribution. Note that in the  intersection geometry shown, a platoon can encounter at most three collision areas to cross the intersection.

\subsubsection{Actions}
To address overlaps on the board, it is essential to delay the entry time of (some) platoons into the collision areas that they intend to cross, recognizing that an optimal solution may involve imposing some delay to some platoons' entry or exit time to some conflict areas (implying a corresponding change to their trajectories). To facilitate this, we use discretised actions to move (or delay) a block up to a preset $N$ moves, i.e., $a \in [1,N]$, and the minimum unit of delay, one move, is 0.1 second. When an action is applied, the corresponding entry and exit time is delayed by $0.1*a$ seconds.

\subsubsection{Reward model}

The reward model is constructed intuitively as given in (\ref{eq:reward}) to facilitate finding shorter crossing schedules

\begin{equation}
\label{eq:reward}
r = \alpha \times \left(1- \frac{t_{\text{cross}}}{T_{max}}\right) + (1-\alpha) \times \left(1- \frac{{step}}{S_{max}}\right)
\end{equation} 

\noindent where we impose an upper limit for the crossing time, denoted as $T_{max}$. Consequently, the more time the final platoon requires to cross, the lower the reward. Additionally, the number of total steps is also taken into account; a larger number of steps required in the solution results in a decreased reward, encouraging a smaller number of delays. The two parts are balanced by a ratio of $\alpha$, ensuring the reward value is between (0,1). Note that when the number of steps in the solution exceed a threshold $S_{max}$, a small negative value of -1e-3 is attributed. In cases where the delay exceeds $T_{max}$, a value of -1 is assigned. In training, we set $T_{max}=30s$, $S_{max}=20$, $N=20$ and $\alpha$ = 0.5.

\subsubsection{Neural network structure}
We introduce a dual-branch neural network architecture specifically designed for board information processing and action selection. The primary branch is responsible for feature extraction from the board state. The network stacks 10 fully connected layers, each of 512 neurons followed by a 1D batch normalisation layer, employing the Exponential Linear Unit (ELU) as the activation function. This branch divides into two specialized heads: the policy head, which outputs a probability distribution over possible actions, and the value head, which approximates the state value for node selection in the tree search. The secondary branch, termed the action mask, operates in parallel and is clipped to $T_{max}$. This branch serves to filter invalid actions based on the input board state. Specifically, valid actions are flagged with a value of 1.0, while invalid actions (moves that result in exceeding $T_{max}$) are assigned a value of $-\inf$. Consequently, after the application of the softmax function, valid actions receive positive probabilities, while invalid actions are zeros.


\subsection{Parallel Neural MCTS}

Unlike classic board games, the initial states in scheduling problems can be highly varied. This variability presents a significant challenge, as training samples can differ substantially, making it difficult for the neural network to 'remember' and stabilize the policy. Another challenge arises when collecting training examples: the tree search can be time-consuming, often taking a long time to identify a solution for a particular trajectory due to the numerous possible outcomes at each step. To strike a balance between performance and efficiency, boundaries are usually set to the tree search process, both by limiting the search depth and by setting a maximum searching time. However, these constraints often result in tree searches struggling to identify solutions along shorter paths. As a result, the training samples are more likely to generate failure signals, potentially causing the policy to become stuck in local optima. To improve the diversity and quality of training data, we transitioned the above framework to a parallel multi-process paradigm. Specifically, each process independently executes an MCTS, initialised with a random state $S$, and employs a copy of the current policy $\pi$. Each MCTS iteration progresses until the end of the actual gameplay. After completion of all the processes, the network undergoes a policy update. This update benefits from the diverse range of initial boards and intermediary state transitions encountered during the parallel search.

\subsection{Curriculum training}
Training follows a curriculum-based approach, specifically designed to adapt the policy from low volume to heavy traffic flow. In high-volume situations, as new platoons approach and trajectory planning begins, the intersection may still be operating under the previous schedule. Our curriculum gradually progresses from a clear intersection to one where preceding schedules are actively crossing,  addressing the challenges posed by the continuous nature of traffic flow. 

\subsubsection{Training on clear intersections}

A 'clear intersection' refers to a low-traffic scenario without any residual, i.e., previously scheduled, platoons crossing the intersection. Tree search algorithms are likely to discover multiple paths from the same initial board state, resulting in trajectories with high variance. To mitigate this, we adopt two modifications. First, we employ gradient accumulation, then average the gradients before performing a policy update in (\ref{eq:crossentropy}). This mitigates the fact that transitions could be cancelled off by each other's gradient contributions. Second, we implement a clipping after gradient accumulation to prevent any unexpectedly large gradients from destabilizing the network update, see (\ref{eq:clip}).

\begin{equation}
\label{eq:crossentropy}
J(\theta)= H(p, q) = -\sum_{x} p(x) \log(q(x))
\end{equation}

\begin{equation}
\label{eq:clip}
J(\theta) = \min \left(\text{clip} \left( \lVert \nabla J(\theta) \rVert_2 , 1 - \epsilon, 1 + \epsilon \right) \right) 
\end{equation}

The objective function \( J(\theta) \) is optimised using the cross-entropy loss to quantify the divergence between the expected distribution derived from the MCTS and the probability distribution predicted by the policy network. Gradient values are clipped within the interval \([1 - \epsilon, 1 + \epsilon]\), where \( \epsilon \) serves as a temperature factor.

\subsubsection{Training on busy intersections}

\begin{figure}[h]
    \centering
    \includegraphics[width=0.4\linewidth]{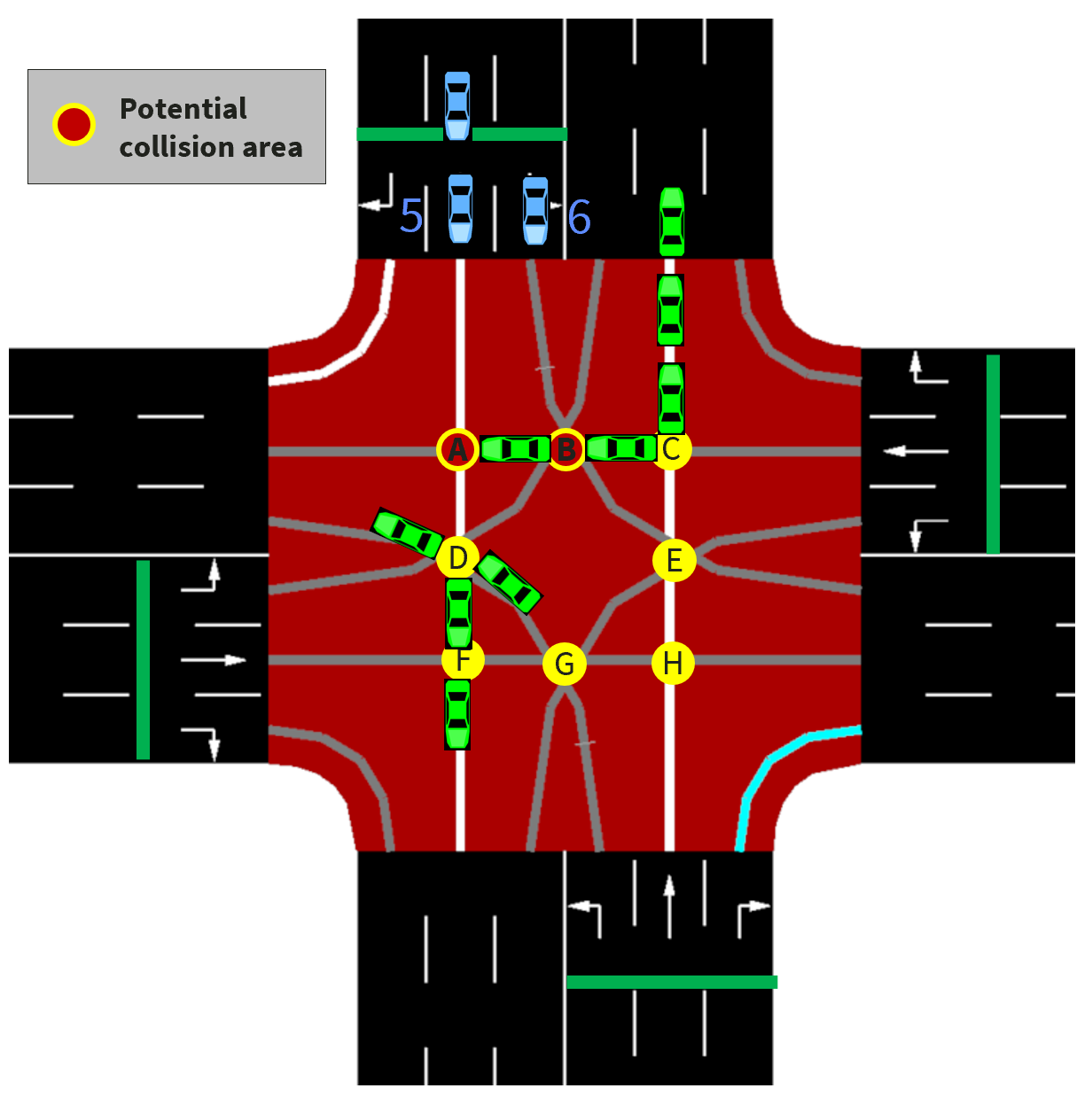}
    \includegraphics[width=0.5\linewidth]{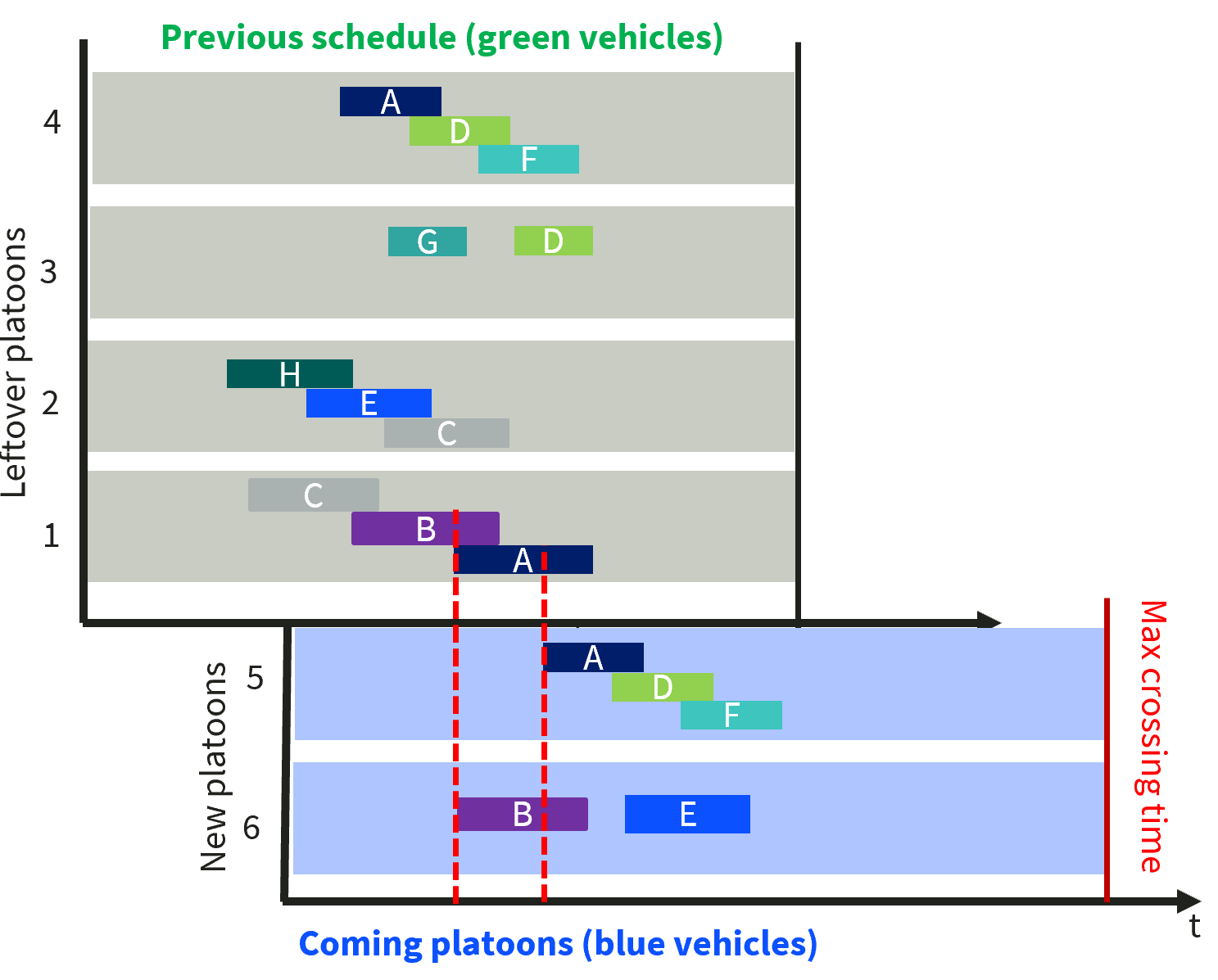} 
   \caption{Transformation board represents a busy intersection that is executing a previous schedule (the upper board) and cannot be rescheduled. The new approaching platoons have potential to collide the previous schedule on collision area A and B. Thus, the current schedule must ensure a compatible solution to the solved board.}
    \label{fig:leftover}
\end{figure}

As the next step in curriculum training, we dynamically simulate complex traffic scenarios by overlaying new traffic boards on to resolved boards randomly sampled from a predefined pool. This simulates the random arrival of new approaching platoons into busy intersections at arbitrary time points. An example of this is given in Fig. \ref{fig:leftover}. The presence of residual platoons can significantly increase the complexity of the situation, potentially doubling the size of the action space. Finding optimal actions via tree search therefore becomes substantially more challenging, introducing more failure trajectories to the memory pool. This complexity can lead to significant performance degradation as the reward signals are indistinguishable from noise, thereby causing the policy to become trapped in local optima.

To mitigate these issues, we employ three key strategies. First, we introduce an entropy regularisation term (\ref{eq:entropyregu}) into the policy objective function to promote exploration and prevent early convergence to suboptimal solutions. The original objective function \( J(\theta) \) can be augmented with an entropy term to encourage exploration. The augmented objective function \( J'(\theta) \) is formulated as (\ref{eq:update}).

\begin{equation}
\label{eq:entropyregu}
H(\pi(a|s; \theta) = - \sum_{i} \pi_{i}(a|s; \theta) \log(\pi_{i}(a|s; \theta))
\end{equation}

\begin{equation}
\label{eq:update}
J'(\theta) = J(\theta) - \beta H(\pi(a|s; \theta))
\end{equation}

\noindent where, \( J(\theta) \) is the original objective function, \( H(\pi(a|s; \theta)) \) is the entropy of the policy \( \pi \) parameterized by \( \theta \) - the higher the entropy, the more uncertain or random the distribution is, and vice versa. \( \beta \) is a hyperparameter controlling the weight of the entropy term. By minimizing \( J'(\theta) \), we effectively maximize the entropy term \( H(\pi(a|s; \theta)) \), provided that \( \beta > 0 \). This encourages the agent to maintain a stochastic policy and promotes exploration, which is particularly useful in environments where the agent has incomplete knowledge. 

Second, we utilize a priority replay mechanism: once a solution for a given board is identified, the best trajectory for that board found to date, termed the 'best-so-far' solution, is archived. This archive is updated continuously throughout the training process, capturing the most effective solutions for each encountered board. We then apply a priority re-sampling strategy during the training:  If a search concludes with a failure signal, the current best-so-far solution is reintroduced into the memory pool with a probability of $\upsilon$ to amplify the associated reward. The probability $\upsilon$ ensures that the best-so-far solution is not always the default recourse,  leaving the policy open to discovering potentially better trajectories.

Last, we incorporate a short-path MCTS roll out into our evaluation methodology in busy intersection scenarios. This method uses a reduced number of simulations and a short tree policy rollout along with the network to select high quality actions. This strategy effectively balances the computational time with yielding reliable decision-making outcomes.

\section{EXPERIMENTS AND ANALYSIS}

\subsection{Experiments}
\subsubsection{Single Intersection Scenarios}

Our initial evaluation objective was to assess the performance of the algorithm on previously encountered boards, i.e., to determine the extent of retained knowledge, specifically, how many solutions from training boards are effectively transferred to the policy, and its efficiency on previously unobserved (zero-shot) boards. To this end, we use a NMCTS policy with an identically structured neural network but initialized with random parameters as our control group. Our experimental group utilizes a neural network that has been fine-tuned.

We then evaluate the performance of learned policies considering two main metrics: success rate and average final reward, with higher values indicating better performance. As a reference point, we employ the FIFO scheduling algorithm as a baseline to compare the scheduled crossing time. A key performance threshold is set at 30 seconds for crossing time; schedules exceeding this limit are considered failures.

To train a PNMCTS-based board solver, which is able to solve busy intersections, we initially train on (simple) synthetic boards representing clear intersections with the four-way, three-lane layout illustrated in Fig. \ref{fig:corresponding} (left) in which only the straight and left-turn lanes cause conflicts (i.e., the right-turn lane is not intersected by other lanes). We generate up to eight platoons with randomly assigned initial parameters (see Table \ref{table: param setup}) per trial.


\begin{table}[h]
\caption{Platoon dynamic setup}
\label{table: param setup}
\label{table_example}
\begin{center}
\begin{tabular}{|c||c||c|}
\hline
\textbf{Parameter} & \textbf{Type} & \textbf{Range}\\
\hline  
Number of platoons& int & [1,8] \\
\hline
Speed & real & [4,5] \\
\hline
Distance to intersection & real & [0,15] \\
\hline
Vehicles per platoon& int & [1,4] \\
\hline
\end{tabular}
\end{center}
\end{table}

We generate 500 unique intersection scenarios that involve collisions, We then apply the proposed transformation model converting the conflict dynamics to initial board states. To test the generality of the proposed model, we split the 500 initial boards into 400 for training and 100 for testing/evaluation.  

We then implement a curriculum learning approach to train the board solver to effectively solve boards representing overlapping (busy) intersections by gradually integrating more complex scheduling tasks.

\subsubsection{Multiple Intersection Scenarios}

We then expand the scope of the study incrementally from a single intersection to a homogeneous 3x3 traffic network. To simulate traffic flow, we uniformly introduce a traffic volume of 600 vehicles/h/road at each entrance to the network using SUMO. The initial control strategy employed is fixed-time traffic light control at all intersections. We progressively increase the proportion of intersections managed by PNMCTS agents, starting from the central intersection and expanding outward. This approach allows us to evaluate the effects on average travel time (ATT) and overall network throughput (TT).

We then conduct experiments under a traffic volume of 700 vehicles/hour/lane to evaluate our method for ATT and TT compared to several state-of-the-art traffic light control methods.

\subsection{Experimental results}

\subsubsection{Single Intersection Scenarios}

\begin{table*}[h]
  \caption{Success rate comparison at synthetic intersections}
  \label{table: clear intersection success rate}
\label{table_example}
\begin{center}
\begin{tabular}{|c||c||c||c||c|}
\hline
    \multicolumn{5}{|c|}{Clear intersection}\\
\hline  
    Policy & Encountered Boards & Success Rate & Unseen Boards & Success Rate  \\
\hline  
    Control policy & 265/400 & 66.25\% & 55/100 & 55.0\%\\
\hline  
    Experimental policy & 385/400 & 96.25\% & 85/100 & 85.0\% \\
\hline
\hline
    \multicolumn{5}{|c|}{Busy intersection}\\
\hline  
    Policy & Encountered Boards & Success Rate & Unseen Boards & Success Rate  \\
\hline 
    Control policy & 120/400 & 30.0\% & 20/100 & 20.0\%\\
    \hline 
    Experimental policy & 360/400 & 90.0\% & 95/100& 95\% \\
\hline
\end{tabular}
\end{center}
\end{table*}

\begin{figure}[h]
    \centering
    \includegraphics[width=0.7\linewidth]{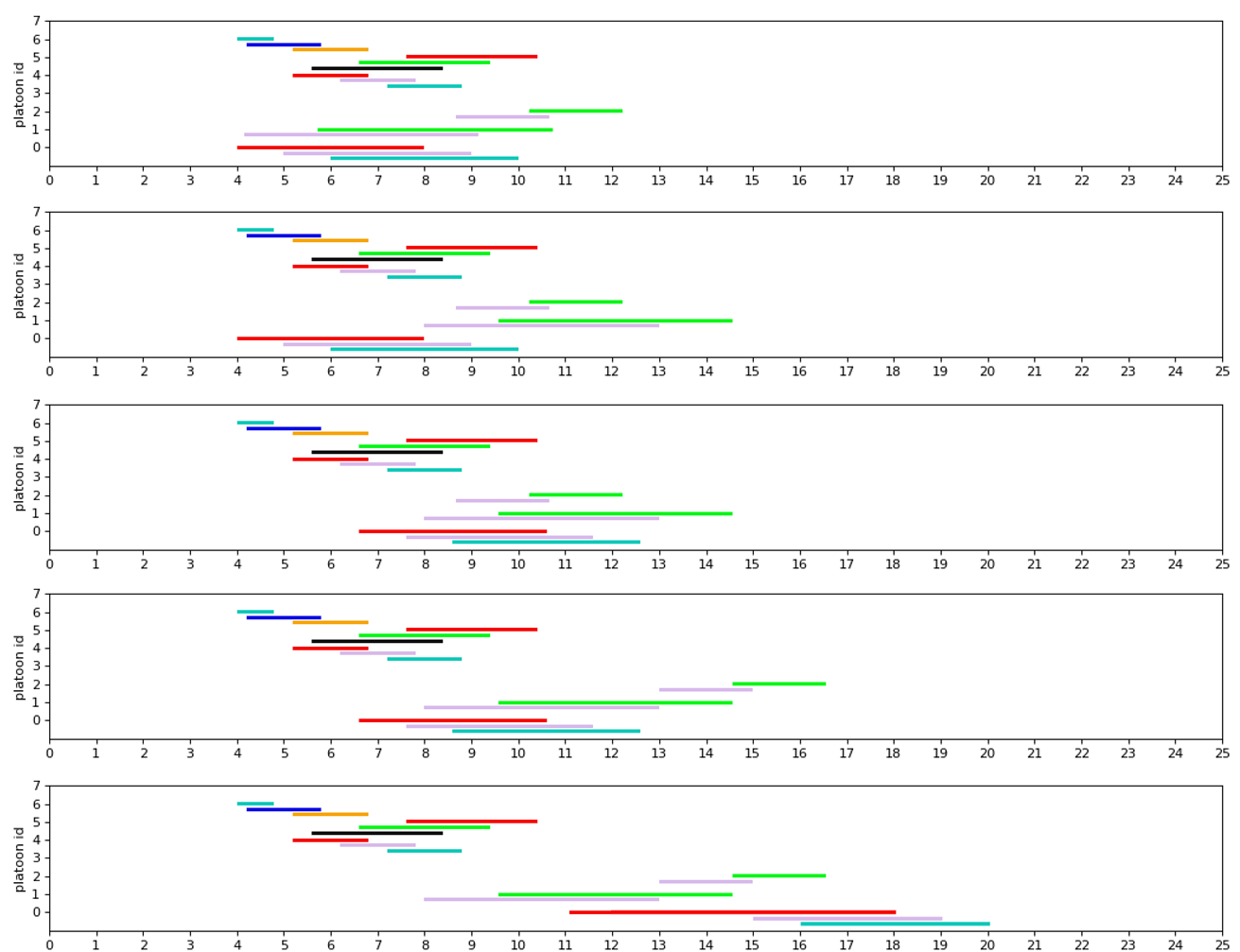}
    \caption{An optimal crossing schedule provided by a PNMCTS agent at a busy intersection from top to bottom, platoon delay sequence: initial board $\rightarrow$ platoon1 $\rightarrow$ platoon0 $\rightarrow$ platoon2 $\rightarrow$ platoon0.}
    \label{fig:step compare}
\end{figure}

The success rates of both the control and experimental policies are shown in Table \ref{table: clear intersection success rate}. The experimental policy performs best on both clear and busy intersections. Importantly, in clear intersection cases, the decisions of the experimental policy were made solely by the policy network. The explanation lies in the network's capability of handling clear intersections, a task that is relatively easy to learn and solve. With  adequate policy network capacity, the experimental policy was capable of solving 96.25\% of training boards and 85\% of unseen boards.

In busy intersection cases, the experimental policy sometimes requires a longer duration to discover optimal solutions, particularly on boards that need multiple steps to solve. The control policy tends to employ a one-step action strategy to reach a suboptimal solution. In contrast, the experimental policy engages in a sequence of actions, executed with finer granularity, to arrive at a better solution (See Fig. \ref{fig:step compare}).


\begin{figure}[h]
    \centering
    \includegraphics[width=0.6\linewidth]{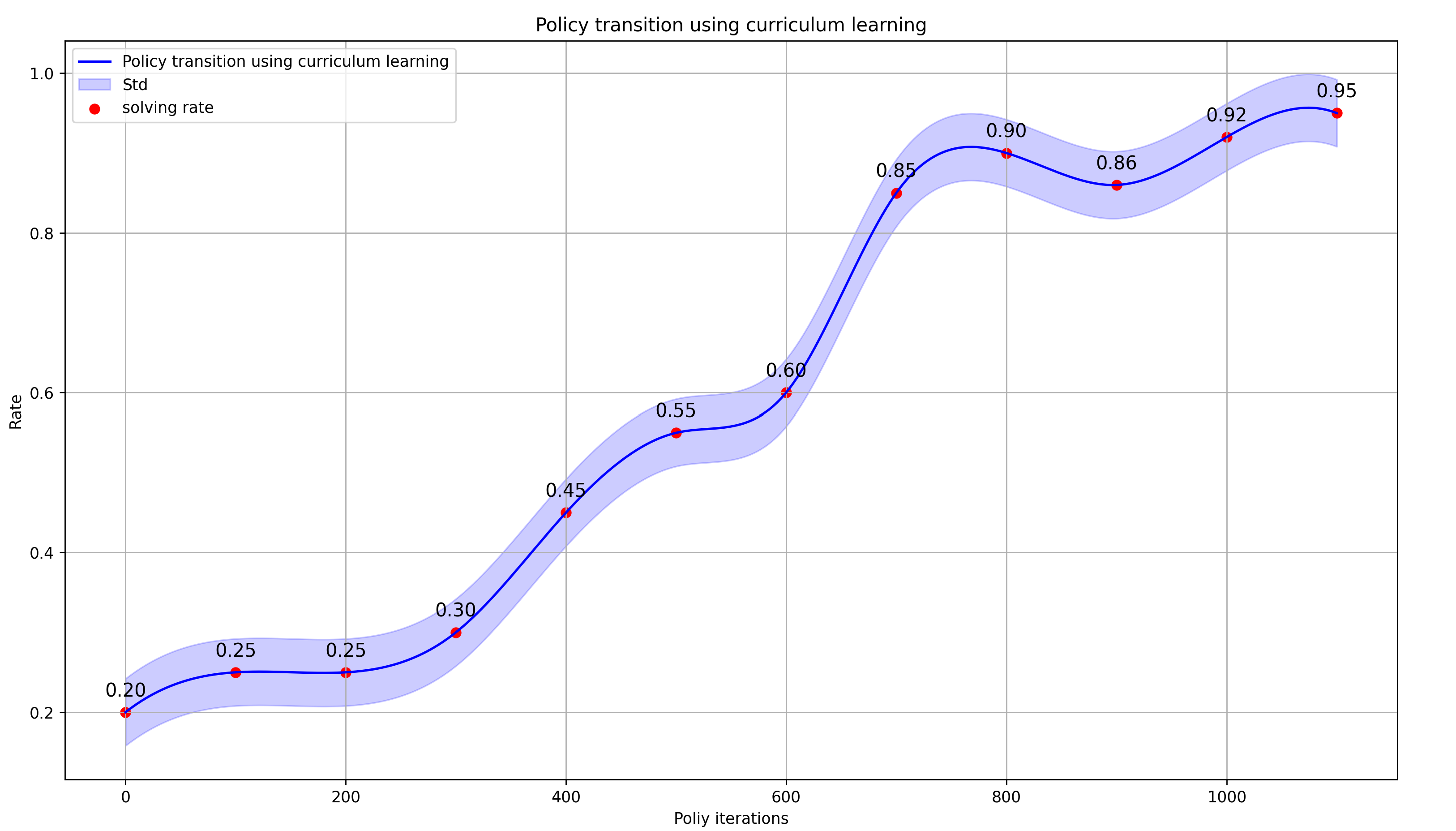} 
    \includegraphics[width=0.6\linewidth]{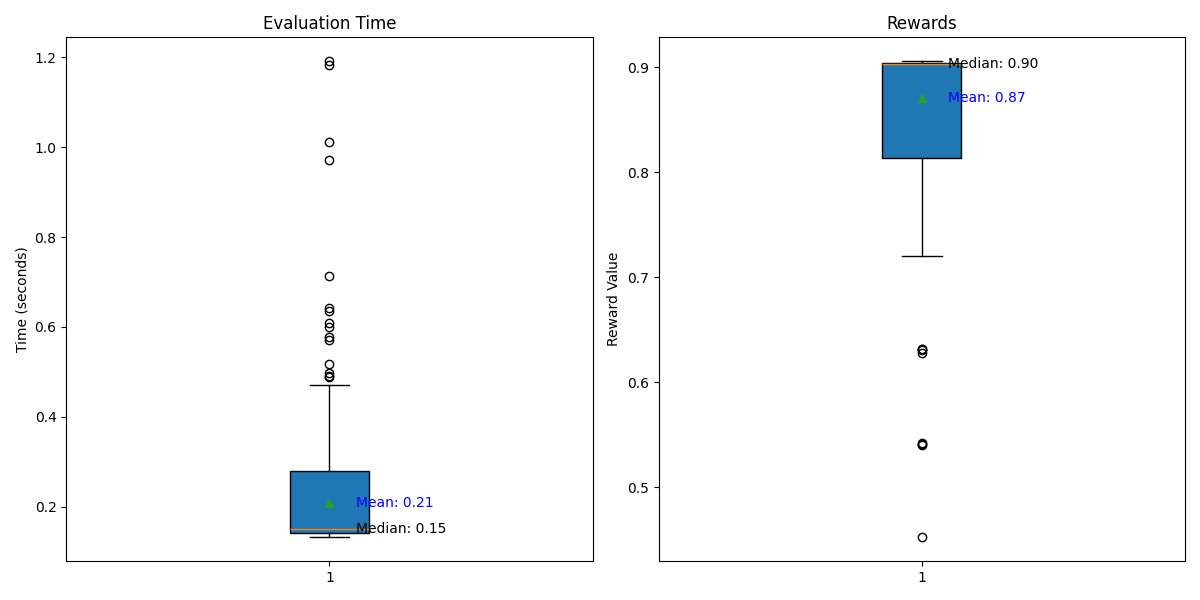} 
    \caption{(a): Performance curve with curriculum training and (b): average board solving time and solution quality distribution.}
    \label{fig:curriculem curve}
\end{figure}

Incorporating a curriculum into the training schedule yields notable performance improvements, transitioning from the control policy to the experimental policy.  The evaluation curve of the curriculum strategy in Fig. \ref{fig:curriculem curve}(a) indicates that as training progresses, a more densely populated solution distribution is observed. On average, the experimental policy solved 95\% of unseen boards, in contrast to the control policy, which managed to solve only 20 boards initially. The policy starts to converge at the 900th policy iteration, the average time spent solving a board drops to 0.21 seconds (median 0.15 seconds), and average solution quality raises to 0.87 (median 0.90) in Fig. \ref{fig:curriculem curve}(b). We can conclude that the experimental policy exhibits good generalisation capabilities on new boards.

\begin{figure}[h]
    \centering
    \includegraphics[width=0.45\linewidth]{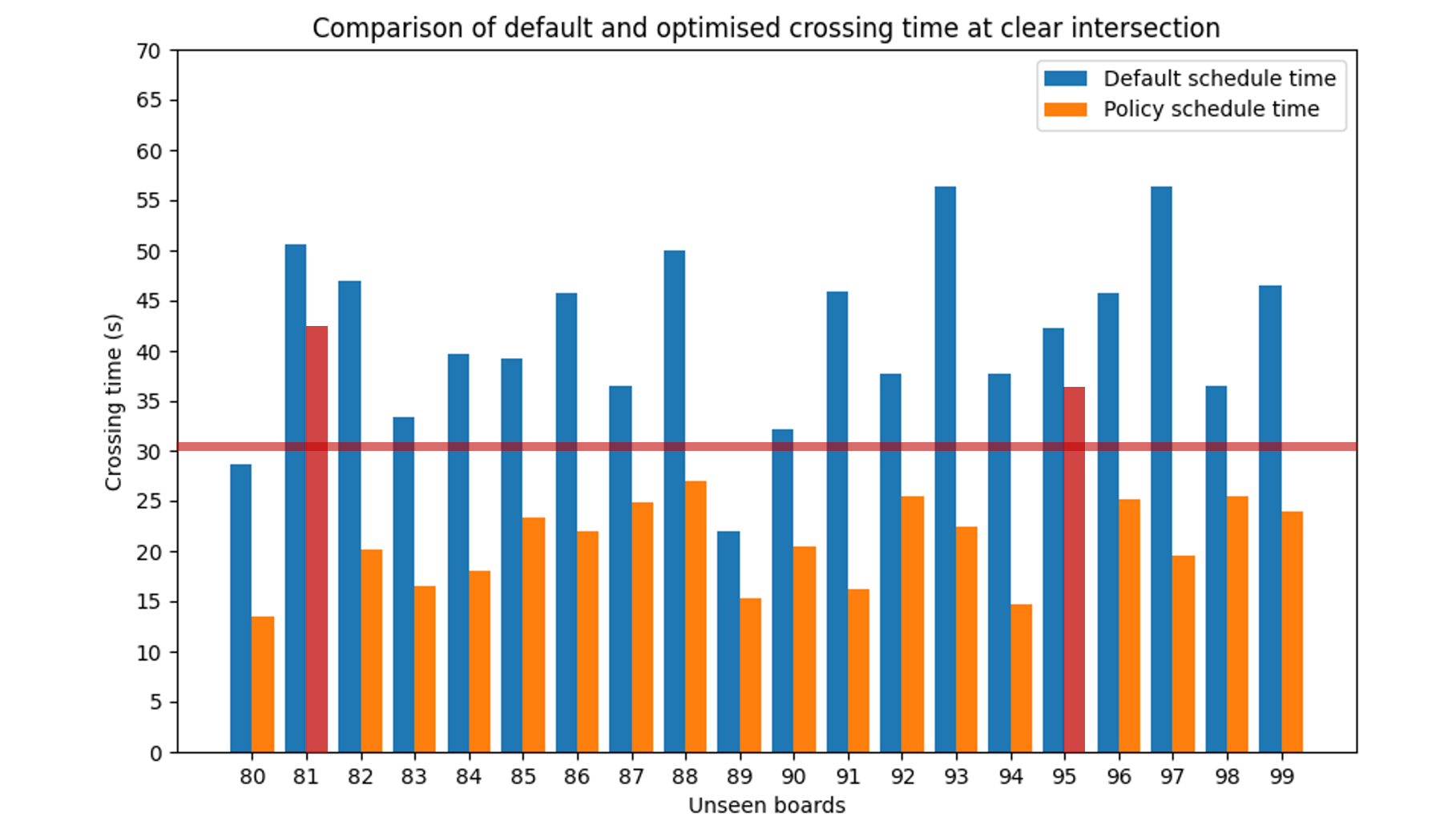}
    \includegraphics[width=0.45\linewidth]{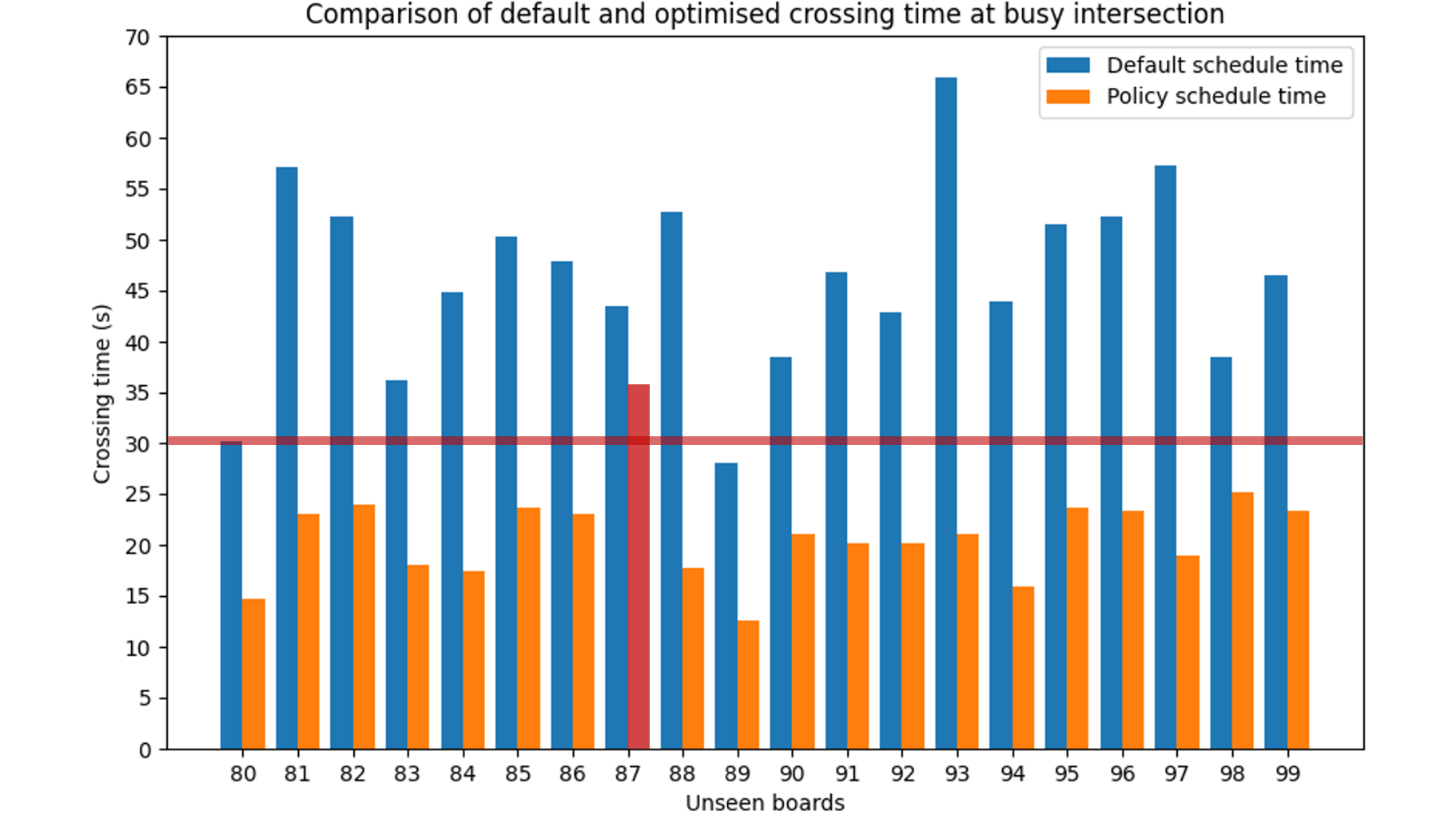} 
    \caption{Crossing time comparison between FIFO and PNMCTS method at, (a). Low-traffic intersections and (b). High-traffic intersections. All solutions provided by PNMCTS outperforms the FIFO in both scenarios.}
    \label{fig:default}
\end{figure}

Next, we compared our method against the FIFO scheduling strategy Fig. \ref{fig:default}. This is a basic traffic control approach where vehicles are allowed to proceed through an intersection in the order which they arrive. This method treats the intersection as a queue, where the first vehicle to arrive at the intersection from any direction has the priority to cross, followed by the next, and so on. In situations where schedules overlap, new incoming platoons are only permitted to enter the intersection after the previous schedule has been fully completed. We present comparison of 20 unseen boards in both clear and busy intersection in Fig. \ref{fig:default}. In low-traffic scenarios Fig. \ref{fig:default}(a), our approach solved 18 cases of schedules completed within 30 seconds and reducing the average crossing time by 43\%. Similarly, in high-traffic scenarios Fig. \ref{fig:default}(b), our method solved 19 cases under the 30-second constraint and reduces the average crossing time by 52\% compared to FIFO. In summary, the results clearly show that our method can identify shorter schedules than the standard FIFO method, in both low and high-traffic situations.

\subsubsection{Multiple Intersection Scenarios}
   
\begin{figure}[h]
    \centering
    \includegraphics[width=0.4\linewidth]{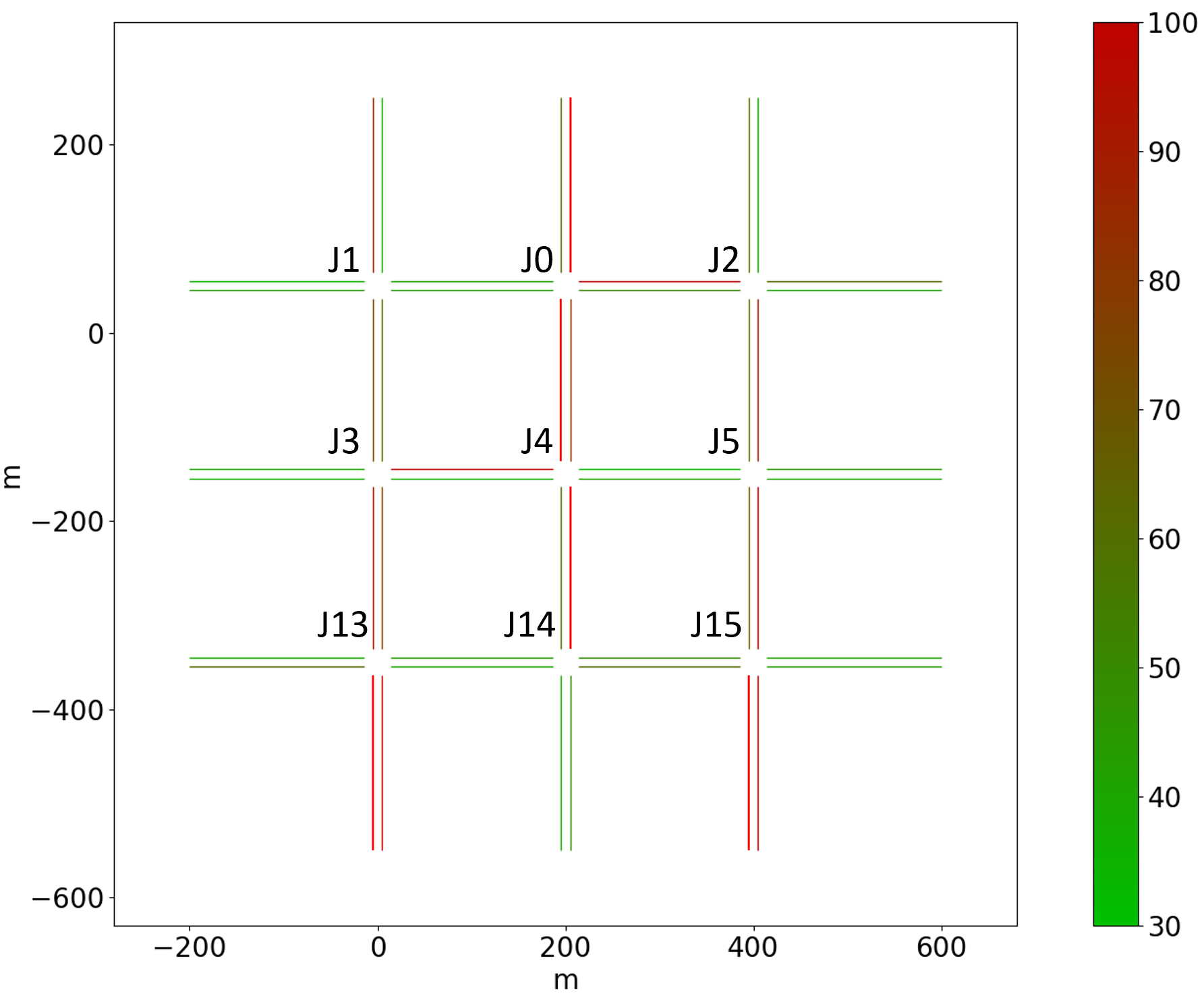}
    \includegraphics[width=0.4\linewidth]{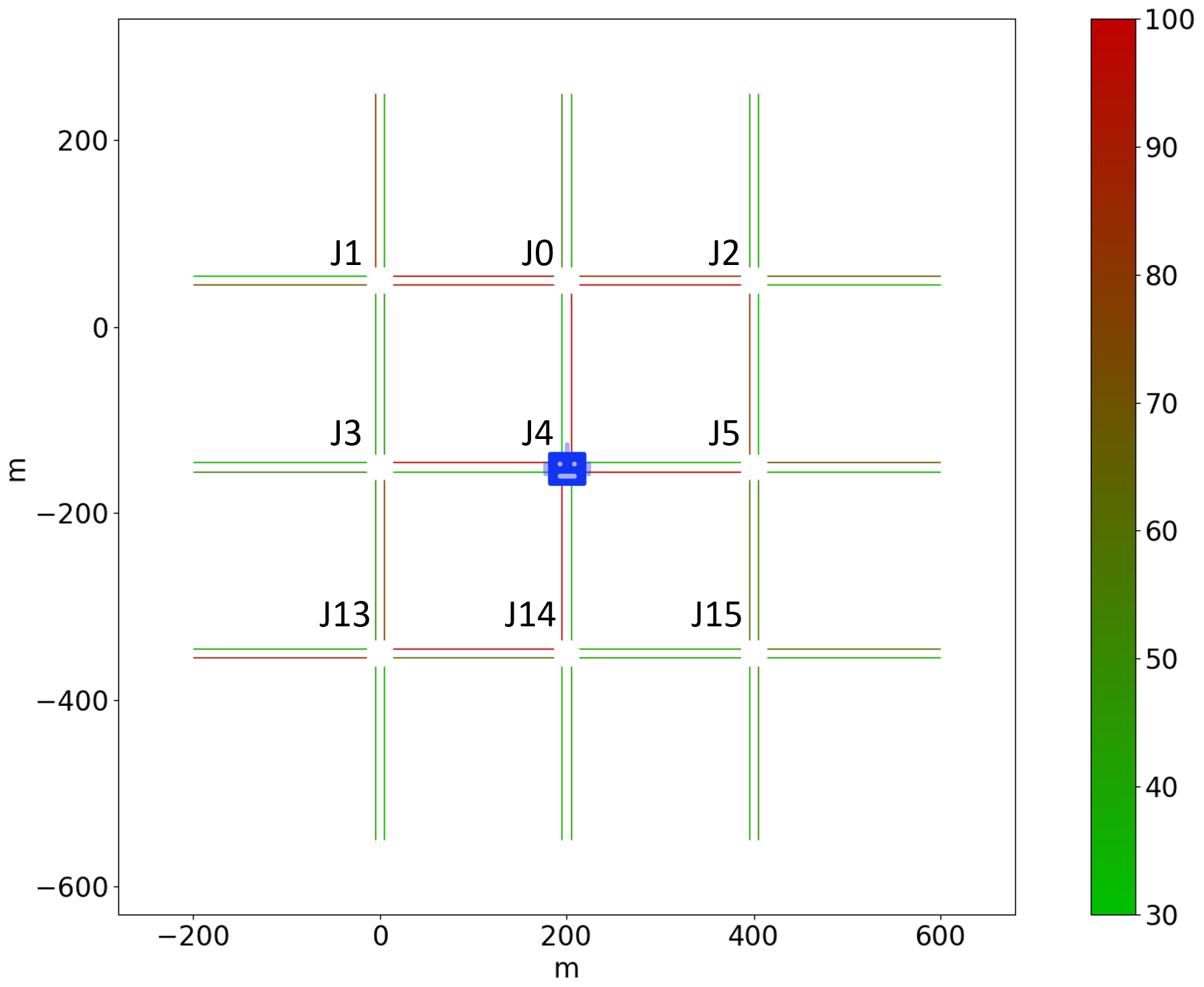}
    \includegraphics[width=0.4\linewidth]{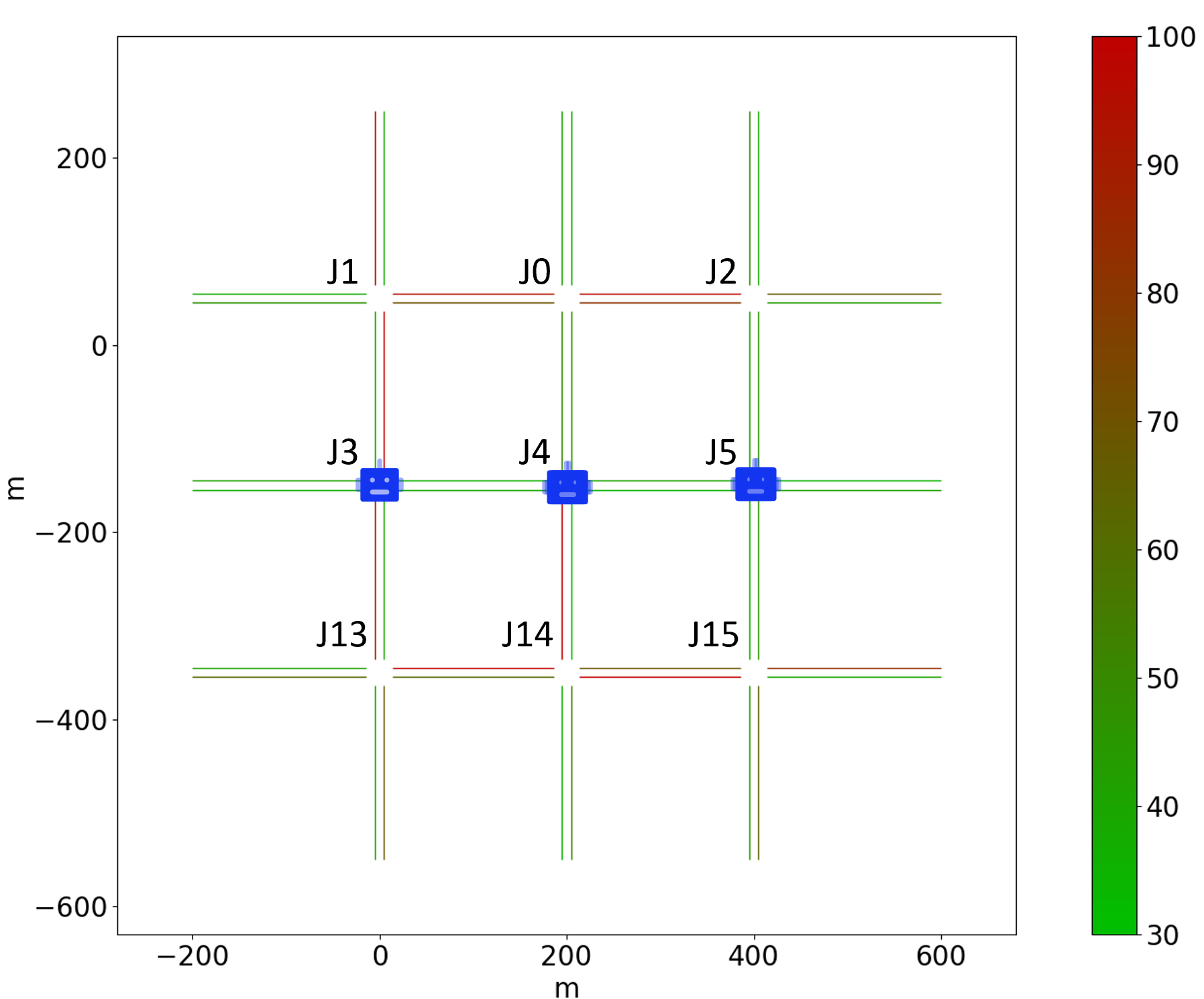} 
    \includegraphics[width=0.4\linewidth]{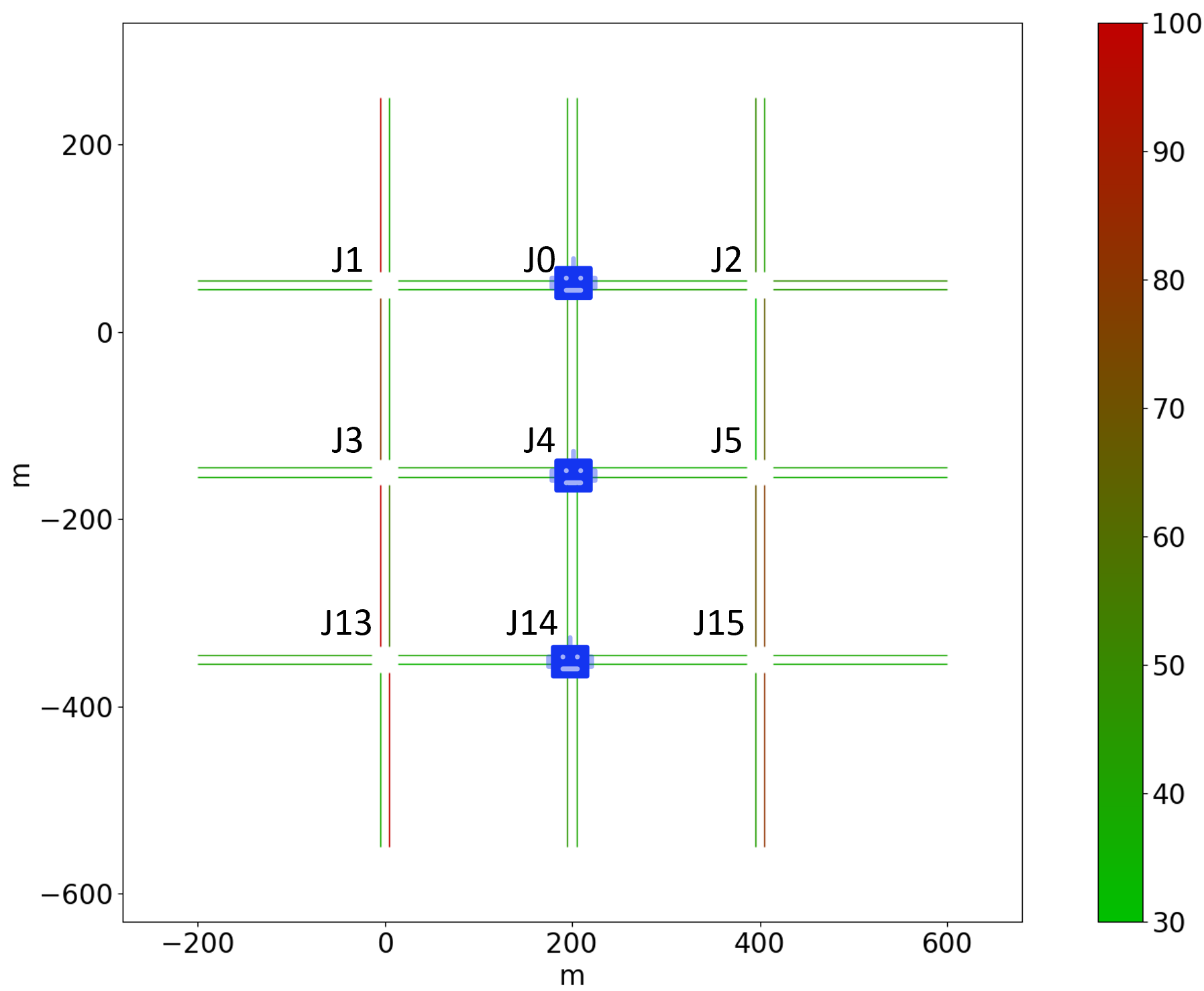}
    \includegraphics[width=0.4\linewidth]{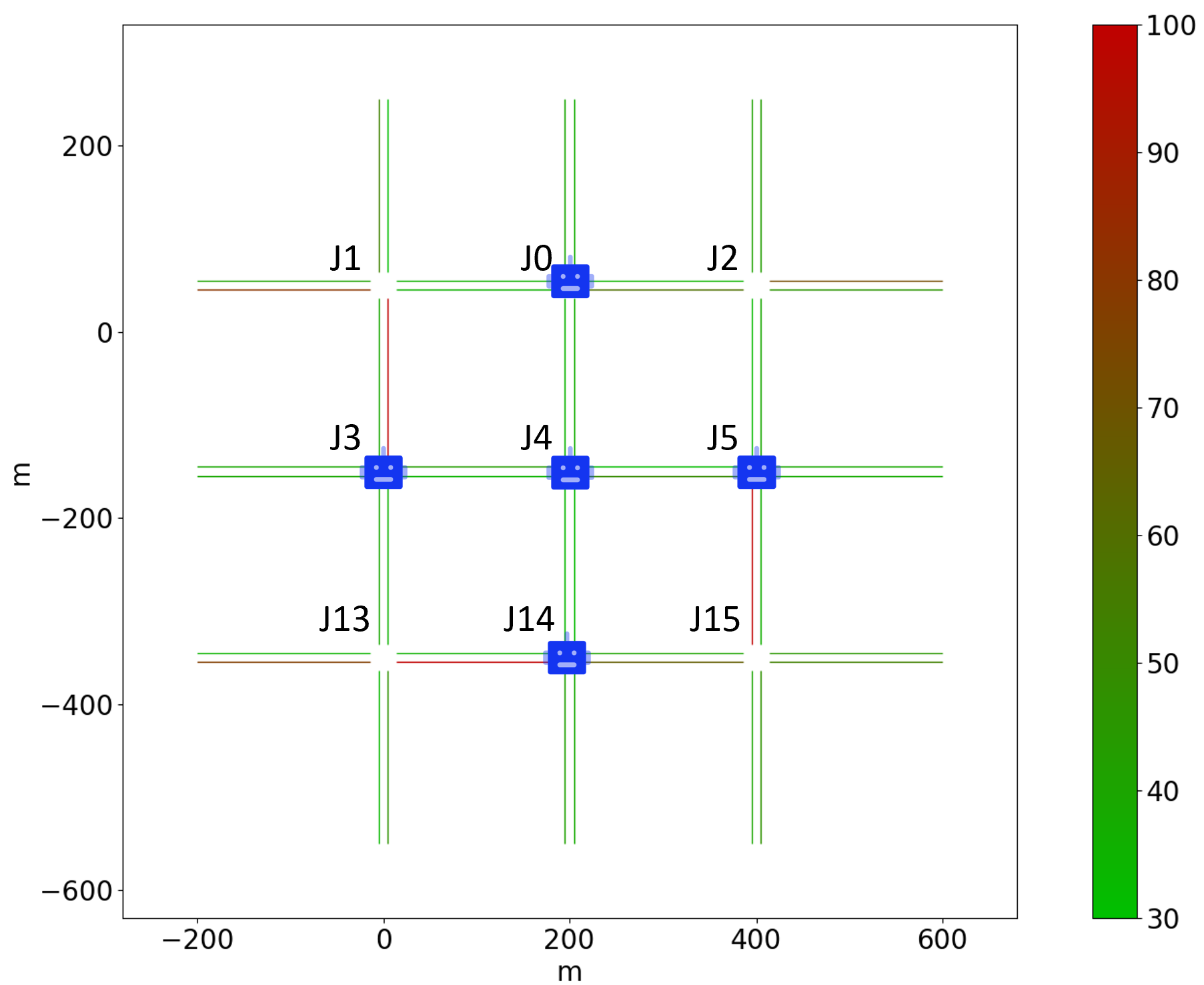} 
    \includegraphics[width=0.4\linewidth]{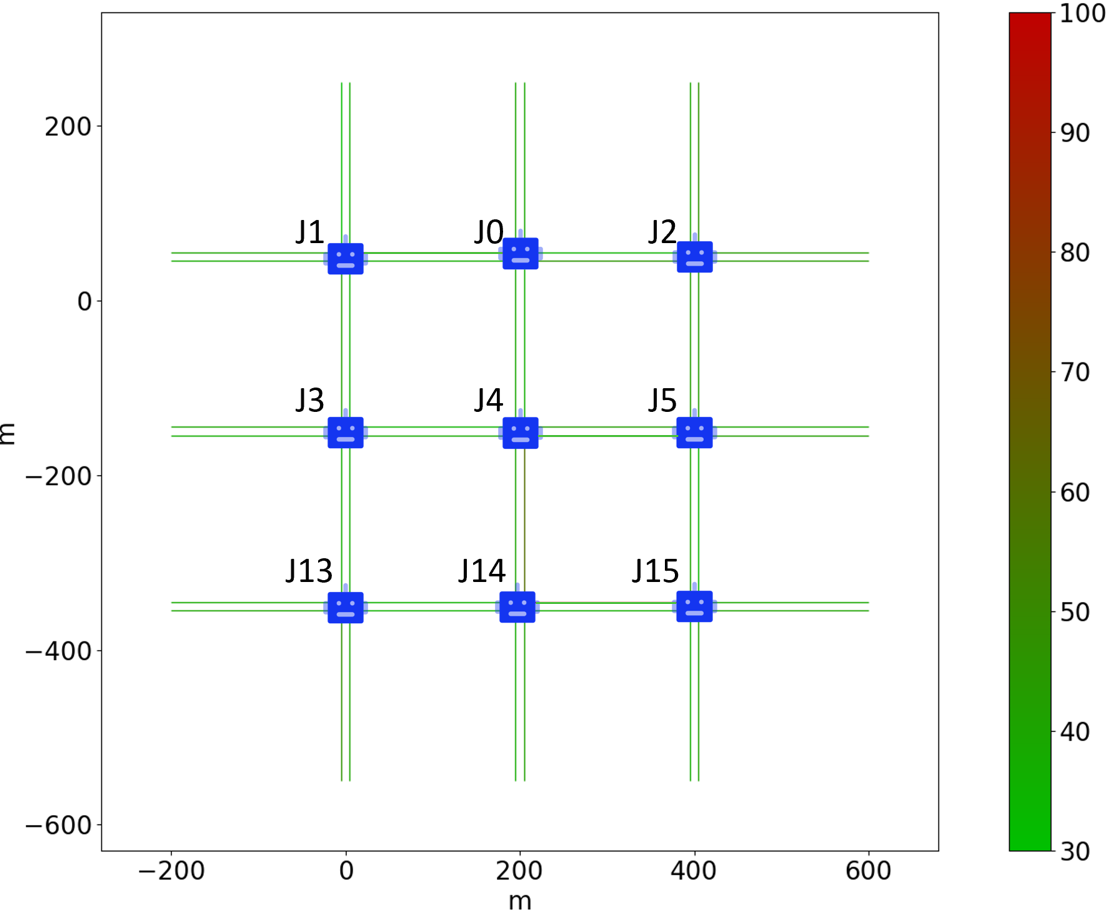} 
    \caption{From left-right top-bottom are scenarios 1-6, respectively. Scenario 1 is default traffic network where all intersections are controlled by fixed-time traffic light controller. When an agent controller is deployed at an intersection, the default fixed-time traffic light controller is replaced. Scenario 6 (right-bottom) is an all agent-controlled intersection network. Controlled by agents, intersections can reduce queue build-up in incoming lanes, as evidenced by the green travel time on upstream roads. A fully agent-controlled network can nearly achieve conditions of free-flow travel.}
    \label{fig:gridtraveltime}
\end{figure}

\begin{figure}[h]
    \centering
    \includegraphics[width=0.45\linewidth]{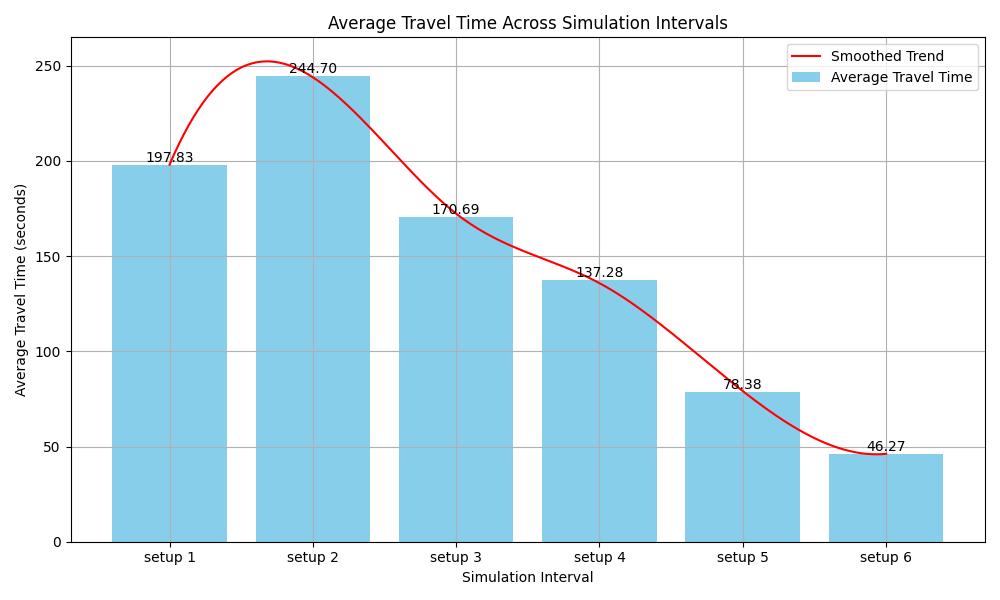}
    \includegraphics[width=0.5\linewidth]{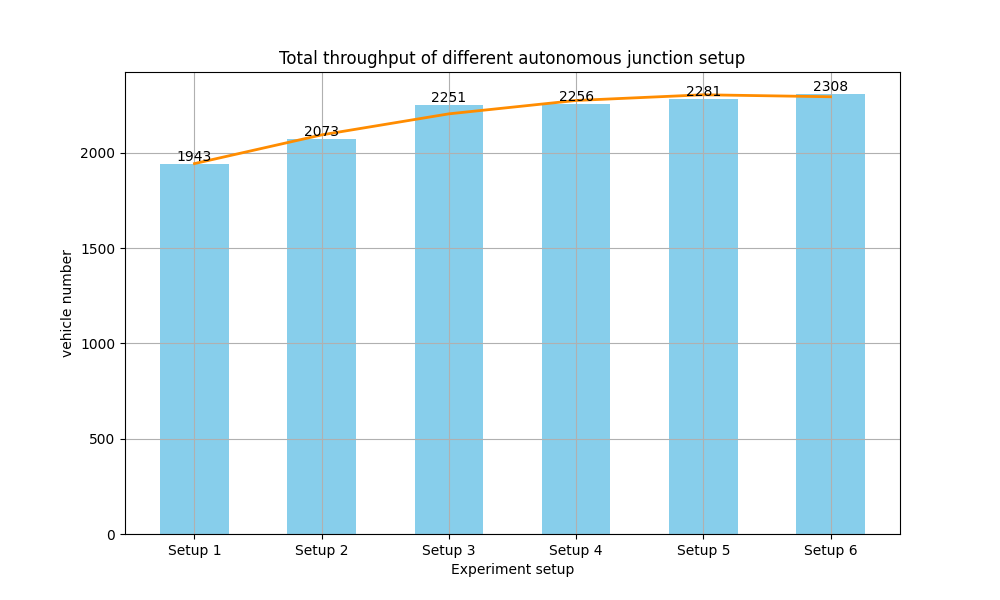}
    \caption{(a): Average travel time and (b): total throughput of the network for scenarios 1-6.}
    \label{fig:trend}
\end{figure}

Initially, the deployment of a single agent-controlled intersection in the network's center (as shown in the second graph in Fig. \ref{fig:gridtraveltime} leads to an increase in average travel time as shown in Fig. \ref{fig:trend}(a). This upturn can be attributed to congestion buildup in the downstream lanes. The presence of an autonomously controlled intersection creates a bottleneck effect, as the traditional traffic control mechanisms surrounding it are unable to efficiently manage the altered traffic patterns.

Subsequent expansion of agent control to additional intersections yields a significant reduction in upstream lane congestion, the greener shading indicating lower ATT, where the presence of all agent-controlled intersections correlates with a free-flow travel condition, as shown in Fig. \ref{fig:trend}(a) with setup 6. Such results suggest that an intelligent control system's ability to reduce queue lengths at intersections directly translates to improved traffic flow and shorter travel time.
\begin{table*}[ht]
\caption{3*3 grid network travel time comparison}
\label{table: performance compare}
\begin{center}
\begin{tabular}{|c||c||c||c||c||c|}
\hline
& \textbf{PNMCTS}&Fixed-time & Actuated & LIT & PressLight \\
\hline  
ATT(s) & \textbf{22.25}& 663.07 & 278.45 & 183.21 & 87.51 \\
\hline
TT(veh) & \textbf{4936}& 1024 & 3203 & 3920 & 4251 \\
\hline
\end{tabular}
\end{center}
\end{table*}
We observed continuous reduction in travel time alongside a plateau in network throughput in Fig. \ref{fig:trend} (b). This can be attributed to the inherent characteristics of traffic flow dynamics and the efficiency of the agent-controlled intersections. As more intersections are managed by intelligent agents, the traffic flow becomes more coordinated, reducing the overall travel time by alleviating bottlenecks and minimizing stop-and-go conditions. This optimized traffic flow allows vehicles to traverse the network without unnecessary stopping, thus decreasing the average travel time across the network. However, the total network throughput, which represents the number of vehicles that pass through the network in a given period, reaches a saturation point. This plateau occurs because there is a maximum carrying capacity within the network—defined by the number of lanes, speed limits, and vehicle dynamics—beyond which no additional improvements in traffic control can increase the number of vehicles that the network can handle. Essentially, while the vehicles that are within the network are moving more rapidly, the overall capacity of the network remains unchanged.

The experimental results validate the efficiency of the proposed PNMCTS method in reducing travel time within a 3x3 grid network. It outperforms traditional fixed-time control, as well as the state-of-the-art methods, such as 'PressLight' reducing travel time by 74.5\% and improving total throughput by 16\% (See Table \ref{table: performance compare}).

In summary, the results from the simulations indicate that the strategic placement and the number of agent-controlled intersections within a traffic network are crucial for optimizing travel time and throughput.

\section{Conclusion}

In conclusion, this study presents a novel application of Parallel Neural Monte Carlo Tree Search (PNMCTS) for addressing the challenges associated with resource scheduling in unsignalized urban traffic management. Our primary contribution is the development of a transformation model that abstracts complex intersection dynamics into a schedulable board game format. By doing so, we recast the problem of managing platoons of vehicles into an NMCTS framework with a clear objective—minimizing intersection crossing time while ensuring a collision-free schedule.

To optimize this framework, we use a parallel version of NMCTS (PNMCTS) that employs four distinct training strategies. First, we leverage parallelism to conduct multiple tree searches and simulations concurrently, each originating from a unique state. Second, we retain and periodically reintroduce optimal trajectories into the training pool to reinforce the reward signal. The combined effect of these strategies is a marked improvement in the quality of the training data. Third, we incorporate entropy regularisation into the policy loss function to encourage a broader exploration strategy, mitigating limitations set by the pruned tree search. Finally, we employ a curriculum learning strategy with a short-path MCTS roll out that enables the policy to progressively adapt to increasingly complex traffic scenarios.

The results from our experimental analysis demonstrate the effectiveness of the PNMCTS method, particularly in managing a four-way unsignalized intersection. PNMCTS has proven capable of handling 95\% of unseen, single busy intersection scenarios, reflecting its robust generalisability. It significantly reduces average crossing time compare to FIFO. Within a homogeneous 3x3 grid network, our strategy maintains free-flow traffic in light traffic and outperforms state-of-the-art RL-based traffic light control in heavy traffic by reducing average travel time by 74.5\% and gain total throughput by 16\%. These results affirm the strength and efficiency of our approach in the domain of intelligent traffic management. Future work will focus on large-scale networks of heterogeneous intersections with dynamic platoon length management.




\section*{ACKNOWLEDGMENT}
 This publication has emanated from research conducted with the financial support of grants from Science Foundation Ireland under Grant number 18/CRT/6223 and SFI Frontiers for the Future grant number 21/FFP-A/8957. For the purpose of Open Access, the authors have applied a CC BY public copyright license to any Author Accepted Manuscript version arising from this submission.

\bibliographystyle{IEEEtran}
\bibliography{root}

\end{document}